\documentclass[conference]{IEEEtran}
\IEEEoverridecommandlockouts
\usepackage{cite}
\usepackage{amsmath,amssymb,amsfonts}
\usepackage{algorithmicx}
\usepackage{algcompatible}
\usepackage{graphicx}
\usepackage{textcomp}
\usepackage{xcolor}
\def\BibTeX{{\rm B\kern-.05em{\sc i\kern-.025em b}\kern-.08em
    T\kern-.1667em\lower.7ex\hbox{E}\kern-.125emX}}

\usepackage{booktabs,array,colortbl,tabularx}
\usepackage{pifont}
\usepackage[bookmarks=false]{hyperref}
\usepackage{booktabs}
\usepackage{microtype}
\usepackage{xspace}
\usepackage{enumitem}
\usepackage{balance}
\usepackage{stfloats}
\usepackage{colortbl}
\usepackage{multicol}
\usepackage{multirow}
\usepackage{pifont}
\usepackage{algorithm}
\usepackage{algpseudocode}
\usepackage{booktabs}
\usepackage{multirow}

\usepackage[table]{xcolor}
\definecolor{tblhead}{HTML}{CBD9E5}
\definecolor{tblsubhead}{HTML}{E4ECF2}
\definecolor{tblbase}{HTML}{F5F6F7}
\definecolor{tblours}{HTML}{E2EFF2}

\begin{document}

\title{\textit{FreqCache}: Frequency-Enhanced Token Caching \\ for Efficient Streaming VLN Models}
\author{
    \IEEEauthorblockN{
        Zihao Zheng$^{1,*}$,
        Hangyu Cao$^{1,*}$,
        Qichao Ma$^{1}$,
        Xichang Wu$^{2}$,
        Zhihao Mao$^{1}$,
        Lingyue Zhang$^{1}$,\\
        Maoliang Li$^{1}$,
        Jiayu Chen$^{1}$,
        Donggang Cao$^{1}$,
        Xuanzhe Liu$^{1}$,
        Hong Mei$^{1}$,
        Xiang Chen$^{1,\dagger}$
    }
    \IEEEauthorblockA{
        $^{1}$ Peking University
        \quad
        $^{2}$ Beijing Jiaotong University\\
        $^{*}$ Equal Contribution
        \quad
        $^{\dagger}$ Corresponding author: xiang.chen@pku.edu.cn
    }
}
\maketitle

\begin{abstract}
Streaming vision-and-language navigation (VLN) repeatedly encodes overlapping views. Robot motion can shift a reusable patch to a different position, while small residual shifts near an edge can change its feature. These effects complicate both patch selection and the choice of a reuse budget for each frame.
We present \textit{FreqCache}, a streaming VLN caching framework that integrates frequency-domain characteristics from raw observations with token-level redundancy identification. 
We analyze their complementarity in cross-frame correspondence, reuse validity, and cache budgeting. Our analysis shows that spectral magnitude and phase cues can recover spatially shifted correspondences, local high-frequency characteristics can reveal structurally sensitive tokens, and single-frame spectral entropy can guide the budget among eligible patches.
Based on these insights, \textit{FreqCache} introduces frequency-enhanced token alignment, selective token refresh, and adaptive cache budgeting to reduce redundant computation while preserving navigation performance.
On R2R-CE, \textit{FreqCache} cuts per-step latency by 6.7--12.0\% relative to VLN-Cache and speeds up inference by up to $1.64\times$ over the original models. Across R2R-CE and RxR-CE, success rate drops by at most 0.7 percentage points.
\end{abstract}
\section{\textbf{Introduction}}
\label{tex:introduction}

\begin{figure}[!t]
    \centering
    \includegraphics[width=3.3in]{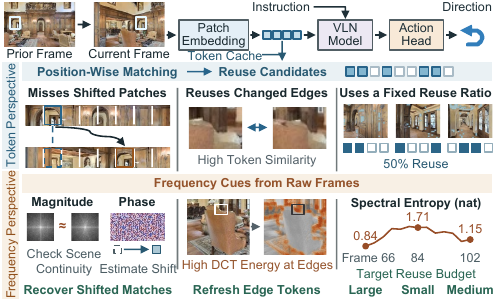}
    \vspace{-2mm}
    \caption{VLN Token Caching Challenges and Frequency Oppotunities}
    \label{fig:1}
    \vspace{-5mm}
\end{figure}

VLN has emerged as a key technical solution for embodied navigation and exhibits strong capabilities in environmental understanding and navigation decision-making~\cite{internvla-n1,chen2024mapgpt}. 
The VLN model temporally encodes observed images and language instructions into tokens, performs reasoning via a large model, and decodes navigation directions through an action head (Fig.~\ref{fig:1}).
Nevertheless, robots must continuously process streaming visual observations and repeatedly perform model inference during continuous motion, incurring substantial computational overhead and hindering real-time deployment~\cite{zhang2024navid,long2024instructnav}.

The temporal continuity of navigation introduces substantial \underline{token/patch redundancy} across consecutive observations, making token caching an effective approach to accelerating streaming VLN through computation reuse~\cite{kang2025harnessing, xu2025vlacache, wei2025streamvln}.
Position-wise matching can miss a previously observed surface when robot motion moves it to another patch~\cite{li2025egoprune,zheng2026vlncache}. View-aligned matching already addresses this displacement~\cite{zheng2026vlncache}. 
Even after alignment, a small residual shift can move a doorframe edge across a patch, changing its feature more than an equally small shift on a smooth wall. We examine what the image spectrum can tell us about both correspondence and the feature error introduced by reuse.

Meanwhile, frequency-domain analysis provides a complementary perspective for characterizing visual redundancy at the signal level.
Unlike token-level representations, frequency characteristics can be directly extracted from raw visual observations, providing complementary cues about visual structures and their temporal dynamics.
For instance, Fourier-VLM~\cite{wang2025fouriervlm} compresses visual tokens using DCT, FreqCa~\cite{liu2025freqca} exploits frequency dynamics for computation reuse in diffusion models, and FAST~\cite{pertsch2025fast} compresses robotic action sequences for efficient representation.
These studies demonstrate the potential of frequency analysis for revealing signal redundancy and its dynamics.
\textit{Motivated by this, we integrate raw-image frequency characteristics with token-level redundancy identification to improve reuse in streaming VLN.}

Token- and frequency-domain representations capture visual redundancy from complementary perspectives, but their different characteristics make effective integration challenging:
\textit{Challenge \ding{172}: Bridging spectral similarity and token correspondence.}
Robot motion shifts visual content across token positions. Frequency characteristics capture scene continuity, but cannot directly establish fine-grained token correspondence.
\textit{Challenge \ding{173}: Reconciling feature similarity and structural changes.}
Token similarity may overlook critical visual changes, while frequency characteristics reveal structural variations but cannot directly determine whether cached tokens remain reusable.
\textit{Challenge \ding{174}: Coordinating global and local redundancy information.}
Frequency characteristics capture overall visual variations, whereas token representations identify local reuse opportunities. Their different scopes complicate consistent caching decisions under dynamic observations.

To address these challenges, we investigate how frequency-domain characteristics complement token-level redundancy from three perspectives.
\textit{Analysis \ding{172}:}
We examine spectral magnitude consistency, phase shifts, and token correspondence under robot motion, revealing how spectral cues complement feature matching to recover shifted token correspondences.
\textit{Analysis \ding{173}:}
We examine local high-frequency responses, token similarity, and reuse-induced feature deviations, revealing how structural cues identify seemingly similar tokens that require refresh.
\textit{Analysis \ding{174}:}
We analyze global spectral dynamics, valid token candidates, and the latency--navigation-performance trade-off, revealing how frequency characteristics and local reuse opportunities determine cache budgets.
Together, these analyses establish the complementary roles of frequency- and token-level redundancy in caching decisions.

Building on these insights, we propose \textit{FreqCache}, a streaming VLN caching framework that exploits frequency- and token-level redundancy through three mechanisms.
\textit{Method \ding{172}:} Frequency-Enhanced Token Alignment combines spectral magnitude consistency and phase correlation with token matching to recover corresponding historical tokens under viewpoint shifts.
\textit{Method \ding{173}:} Frequency-Enhanced Token Refresh integrates local DCT-based high-frequency energy with token similarity to identify structurally sensitive tokens for selective recomputation.
\textit{Method \ding{174}:} Frequency-Token-Aware Cache Budgeting combines global spectral entropy with token candidates to adapt reuse budgets to scene dynamics while restricting reuse to eligible tokens.

Our contributions are threefold.
\begin{itemize}
    \item We characterize when position-wise token matching misses reusable patches and when similar patches incur different reuse errors.
    \item We implement frequency-guided alignment, refresh, and budgeting on the GPU while preserving visual sequence length and patch order.
    \item On R2R-CE with A100 and Jetson Orin, \textit{FreqCache} accelerates DualVLN by $1.63$--$1.64\times$ and StreamVLN by $1.43$--$1.44\times$. Across R2R-CE and RxR-CE, SR and SPL each drop by at most 0.7 percentage points.
\end{itemize}
\section{\textbf{Background and Motivation}}
\label{tex:background}

\subsection{\textbf{Token Caching for Streaming VLN Inference}}
\label{tex:background_1}

\subsubsection{\textbf{Streaming VLN Models}}
Streaming VLN models continuously process visual observations and language instructions to generate navigation actions.
Recent approaches leverage video-based modeling and streaming context management to support long-horizon navigation~\cite{zhang2025uninavid, wei2025streamvln}.
However, repeated inference over high-resolution visual inputs introduces substantial computational overhead.

\subsubsection{\textbf{Token Redundancy during VLN Inference}}
Due to the temporal continuity of navigation, consecutive observations often share overlapping scene content, leading to redundant token computations.
Recent studies exploit such temporal redundancy by reusing historical visual representations to reduce repeated processing~\cite{kang2025harnessing, xu2025vlacache}.
However, robot motion continuously changes the spatial distribution of visual information, complicating computation reuse.

\subsubsection{\textbf{Token Caching for VLN Inference Acceleration}}
VLA-Cache~\cite{xu2025vlacache} reuses visual KV states and adapts reuse across layers. VLN-Cache~\cite{zheng2026vlncache} aligns patches using depth and relative pose, filters them by task relevance, and sets layer budgets from attention entropy. \textit{FreqCache} uses image phase for alignment, local DCT energy for refresh, and image spectral entropy for a frame-level budget.

\subsection{\textbf{Frequency-Based Model Optimization}}
\label{tex:background_2}

\subsubsection{\textbf{Frequency-Based Model Redundancy Analysis}}
Recent studies exploit frequency-domain characteristics for efficient model inference. 
Fourier-VLM~\cite{wang2025fouriervlm} compresses visual tokens via DCT, FreqTS~\cite{yang2025freqts} and FreqCa~\cite{liu2025freqca} leverage spectral features for token selection and computation reuse in diffusion models, while recent works extend frequency-guided compression to LLM and multimodal KV caches~\cite{kai2026freqkv, yang2026flashcache}. 
These studies highlight the potential of integrating frequency- and token-level redundancy for more effective computation reuse and efficient inference.

\subsubsection{\textbf{Challenges of Frequency-Guided VLN Token Caching}}
Frequency- and token-level representations capture complementary aspects of visual redundancy, yet integrating them into consistent caching decisions remains challenging.
Specifically, 1) spectral similarity and token correspondence are difficult to reconcile under robot motion;
2) feature similarity and structural changes provide different cues for determining reuse validity;
and 3) global scene variations and local token reuse opportunities are difficult to coordinate for cache budgeting.
Addressing these challenges requires jointly considering frequency- and token-level information in cross-frame alignment, selective refresh, and adaptive budgeting.
\section{\textbf{Frequency- and Token-Level Redundancy Analysis for Streaming VLN}}
\label{tex:analysis}

Token matching identifies reuse opportunities, while frequency cues from raw observations capture scene continuity and structural changes. We investigate how they support cross-frame correspondence, reuse validity, and adaptive cache budgeting.

\vspace{-1mm}
\subsection{\textbf{Frequency-Token Cross-Frame Alignment}}
\label{tex:analysis_1}

\subsubsection{\textbf{Limitation of Position-Wise Token Matching}}
Token-level matching typically compares patches at the same spatial positions across consecutive observations. However, robot motion can shift the same scene content to different patch positions, causing position-wise matching to overlook reusable historical features. We characterize this limitation using spatial token similarity (Eq.~\eqref{eq:spatial_similarity}), where $p^t_{i,j}$ denotes the patch at position $(i,j)$, $\operatorname{emb}(\cdot)$ its token representation, and $\mathcal{O}$ the set of compared positions. In Fig.~\ref{fig:freq_alignment}, same-position lookup returns a windowsill, while the cached chair back lies 29 pixels to the left.
\begin{equation}
\operatorname{Sim}_{\mathrm{spa}}(t,t+1)
= \frac{1}{|\mathcal{O}|}\sum_{(i,j)\in\mathcal{O}}
\frac{\operatorname{emb}(p^{t}_{i,j})^{\top}
\operatorname{emb}(p^{t+1}_{i,j})}
{\|\operatorname{emb}(p^{t}_{i,j})\|_2\,
 \|\operatorname{emb}(p^{t+1}_{i,j})\|_2},
\label{eq:spatial_similarity}
\end{equation}

\subsubsection{\textbf{Frequency Cues for Recovering Shifted Correspondence}}
\begin{figure}[!t]
    \centering
    \includegraphics[width=3.3in]{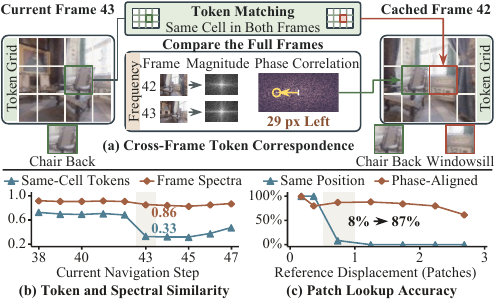}
    \vspace{-2mm}
    \caption{Magnitude Spectra Remain Similar across Shifted Views, and Phase Correlation Recovers Patches Missed by Same-Position Matching.}
    \label{fig:freq_alignment}
    \vspace{-4mm}
\end{figure}

To complement position-wise token matching, we analyze frequency characteristics directly from raw observations. 
According to the Fourier Shift Theorem, spatial translation preserves spectral magnitude while inducing phase changes. We measure cross-frame magnitude consistency as:
\begin{equation}
\operatorname{Sim}_{\mathrm{freq}}(t,t+1)
=\frac{\langle A_t,A_{t+1}\rangle}
{\|A_t\|_2\,\|A_{t+1}\|_2},
\label{eq:frequency_similarity}
\end{equation}
where $A_t$ denotes the magnitude spectrum of observation $t$. 
As shown in Fig.~\ref{fig:freq_alignment}, at step 43, same-position token similarity falls to 0.33 while spectral magnitude similarity remains at 0.86 (Fig.~\ref{fig:freq_alignment}~(b)). In the highlighted displacement range, phase alignment raises patch lookup accuracy from 8\% to 87\% (Fig.~\ref{fig:freq_alignment}~(c)). The phase peak estimates one dominant image-plane translation. It can recover a shifted patch within the overlapping view, as in Fig.~\ref{fig:freq_alignment}(a). Rotation and depth-dependent parallax can leave different residual displacements across patches, which motivates checking individual candidates after alignment.

\noindent \textit{\textbf{Insight \ding{172}:} Spectral magnitude reveals shared scene content beyond position-wise token matching, while phase cues help recover corresponding historical tokens under spatial shifts.}

\vspace{-1mm}
\subsection{\textbf{Assessing Reuse Validity with Local Spectral Cues}}
\label{tex:analysis_2}

\subsubsection{\textbf{Token Similarity Overlooks Structural Changes}}
Cross-frame correspondence identifies potential reuse candidates, but token similarity does not necessarily guarantee reuse validity. Structural changes around obstacle boundaries, door frames, or corners may be overlooked by feature matching, even when corresponding tokens remain highly similar. Reusing these outdated features can introduce errors into subsequent visual representations. We therefore investigate whether frequency characteristics can reveal such changes and complement token-level reuse assessment.

\subsubsection{\textbf{Relating High-Frequency Energy to Reuse Errors}}
\begin{figure}[!t]
    \centering
    \includegraphics[width=3.3in]{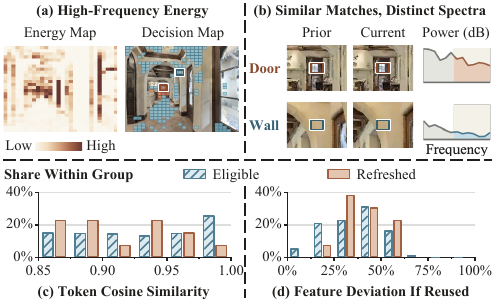}
    \vspace{-2mm}
    \caption{Local DCT Energy and Feature Deviation under Reuse. Histograms Are Normalized within the Eligible and Refreshed Groups.}
    \label{fig:local_sensitivity}
    \vspace{-4mm}
\end{figure}

High-frequency components respond strongly to edges and fine-grained details. We measure patch-wise DCT high-frequency energy as Eq.~\eqref{eq:high_frequency_energy}, where $D_{p_{i,j}}$ denotes the DCT coefficients of a $P\times P$ image patch and $H$ is a high-pass mask. 
\begin{equation}
E_{i,j}=\sum_{u=0}^{P-1}\sum_{v=0}^{P-1}
\left(H(u,v)\,D_{p_{i,j}}(u,v)\right)^2,
\label{eq:high_frequency_energy}
\end{equation}
To examine how this structural cue relates to the consequences of reusing historical features, we measure the deviation between cached and fully recomputed token representations:
\begin{equation}
D_{\mathrm{feat}}^{i,j}=
\frac{\|F_{\mathrm{cache}}^{i,j}-F_{\mathrm{full}}^{i,j}\|_2}
{\|F_{\mathrm{full}}^{i,j}\|_2+\epsilon}.
\label{eq:feature_deviation}
\end{equation}
The door and wall examples in Fig.~\ref{fig:local_sensitivity}(b) have different spectra despite similar cross-frame matches. The eligible and refreshed groups overlap in cosine similarity (c), yet a larger fraction of refreshed patches exceeds 25\% feature deviation under reuse (d). We measure $D_{\mathrm{feat}}$ offline. Removing high-frequency cues lowers SPL by 0.7 points on each model (Table~\ref{tab:freqcache_ablation}).

\noindent \textit{\textbf{Insight \ding{173}:} Local frequency characteristics complement token similarity with structural sensitivity, helping identify potentially unsafe reuse candidates for selective refresh.}

\begin{figure}[!b]
    \centering
    \vspace{-4mm}
    \includegraphics[width=3.3in]{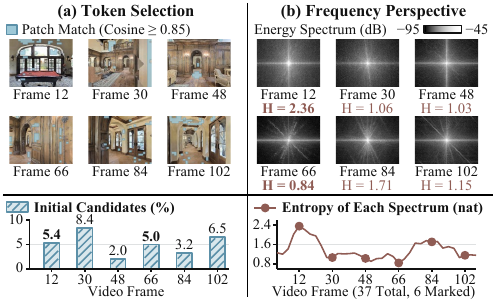}
    \vspace{-2mm}
    \caption{Initial Patch Match Rates for Six Views (a) and Spectral Entropy across 37 Sampled Frames, with the Same Six Views Marked (b).}
    \label{fig:spectral_entropy}
\end{figure}

\subsection{\textbf{Frequency-Adaptive Budgeting Beyond Token Eligibility}}
\label{tex:analysis_3}

\subsubsection{\textbf{Token Eligibility under Varying Scene Complexity}}
The initial match rate measures how often a patch resembles the patch at the same position in the previous frame. Spectral entropy measures how power is distributed across frequencies within the current frame. Changing the previous frame can change the match rate even when the current frame, and hence its entropy, stays fixed. We compare these two quantities in Fig.~\ref{fig:spectral_entropy} before using entropy to set a target and the eligible set to bound actual reuse.

\subsubsection{\textbf{Spectral Entropy as a Cue for Scene Complexity}}
To characterize visual complexity beyond token-level reuse eligibility, we measure the dispersion of spectral energy in each raw observation using spectral entropy:
\begin{equation}
\Psi_t=-\sum_{u,v}
\frac{A_t(u,v)^2}{\|A_t\|_F^2}
\log\frac{A_t(u,v)^2}{\|A_t\|_F^2}.
\label{eq:spectral_entropy}
\end{equation}
Lower entropy indicates more concentrated spectral energy, whereas higher entropy reflects a more dispersed frequency distribution, often associated with richer visual structures and details. 
In Fig.~\ref{fig:spectral_entropy}~(a), frames 12 and 66 have similar initial candidate ratios (5.4\% and 5.0\%), yet their spectral entropies differ markedly (2.36 and 0.84 nats, respectively) in Fig.~\ref{fig:spectral_entropy}~(b).
Conversely, frames 30 and 48 have nearly identical entropies (1.06 and 1.03 nats) but different candidate ratios (8.4\% and 2.0\%). This motivates using entropy to set the target reuse budget and the number of eligible tokens to cap actual reuse.

\noindent \textit{\textbf{Insight \ding{174}:} Token-level selection identifies eligible reuse candidates, while spectral entropy characterizes visual complexity to dynamically adjust how many tokens to reuse.}
\section{\textbf{\textit{FreqCache} Framework: Frequency--Token Fusion for VLN Caching}}
\label{tex:method}

\begin{figure*}[!t]
    \centering
    \includegraphics[width=6.9in]{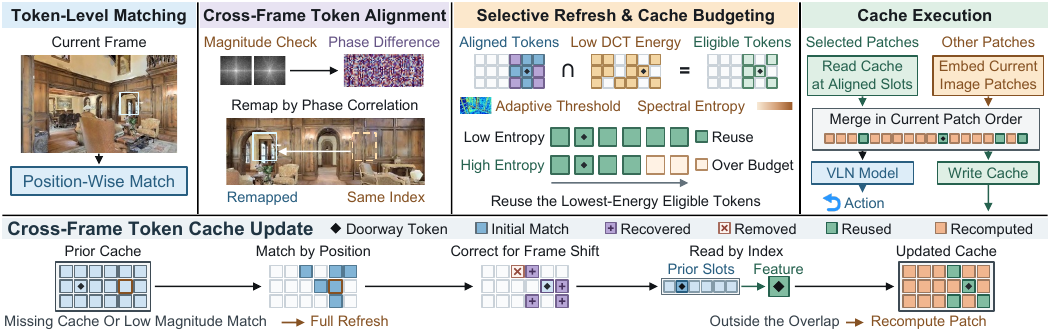}
    \vspace{-2mm}
    \caption{Frequency-Guided Token Alignment, Selective Refresh, and Adaptive Cache Budgeting in \textit{FreqCache} Framework}
    \label{fig:freqcache}
    \vspace{-4mm}
\end{figure*}

Fig.~\ref{fig:freqcache} shows the cache framework. A missing cache or frame history triggers a full refresh, as does magnitude similarity below $\tau_{\mathrm{mig}}$. Otherwise, phase correlation refines patch matches, local DCT energy filters candidates, and spectral entropy sets the target reuse fraction.

\vspace{-1mm}
\subsection{\textbf{Frequency-Enhanced Token Alignment}}
\label{tex:method_1}

\subsubsection{\textbf{Token-Level Matching for Initial Candidate Screening}}
We first perform position-wise matching between current patch tokens and historical tokens to form preliminary reuse candidates $\mathcal{P}_t$. These matches provide an initial correspondence, but may miss shared scene content shifted across patch positions by robot motion.

\subsubsection{\textbf{Magnitude Validation and Phase-Guided Correspondence Refinement}}
We test cross-frame magnitude similarity using Eq.~\eqref{eq:frequency_similarity}. A value below $\tau_{\mathrm{mig}}$ triggers a full refresh and cache reset. Otherwise, phase correlation estimates the dominant image displacement, which we map to the patch grid. Because one displacement represents the frame pair, patches on surfaces at different depths can retain different alignment errors. Within the overlap, a patch that missed at its original index can recover a match at a displaced index (Fig.~\ref{fig:freqcache}). For each accepted current position $p\in\mathcal{A}_t$, we retain its historical source index $\pi_t(p)$. This index retrieves the cached patch embedding, while $p$ specifies its destination in the current frame. We recompute patches outside the overlap.

\vspace{-1mm}
\subsection{\textbf{Frequency-Enhanced Token Refresh}}
\label{tex:method_2}

\subsubsection{\textbf{Aligned Token Candidates for Frequency-Based Refresh}}
The aligned set $\mathcal{A}_t$ provides historical features for reuse assessment. However, correspondence alone cannot guarantee validity when structures around edges, corners, or obstacle boundaries change. We therefore use frequency characteristics to further filter the aligned candidates.

\subsubsection{\textbf{DCT High-Frequency Filtering for Selective Refresh}}
We compute patch-wise DCT high-frequency energy $E_{i,j}$ using Eq.~\eqref{eq:high_frequency_energy}, and set an adaptive threshold:
\begin{equation}
\theta_E^t=\mu_E^t+\lambda\sigma_E^t,
\label{eq:refresh_threshold}
\end{equation}
where $\mu_E^t$ and $\sigma_E^t$ are the mean and standard deviation of patch energies. Aligned tokens with energy above $\theta_E^t$ are refreshed, while the remaining tokens form the eligible set $\mathcal{C}_t$. Token matching therefore establishes the reuse candidates, and DCT energy excludes structurally sensitive tokens. The feature deviation $D_{\mathrm{feat}}$ in Sec.~\ref{tex:analysis_2} is used only for offline analysis, not inference.

\subsection{\textbf{Frequency--Token-Aware Cache Budgeting}}
\label{tex:method_3}

\subsubsection{\textbf{Spectral-Entropy-Based Adaptation of the Token Reuse Budget}}
The eligible set $\mathcal{C}_t$ specifies which historical tokens can be reused, while spectral entropy $\Psi_t$ (Eq.~\eqref{eq:spectral_entropy}) estimates the visual complexity of the current observation. Candidate availability alone does not capture the scene-dependent reuse demand. We combine both sources to determine the reuse budget:
\begin{equation}
\begin{aligned}
\alpha_t &= \alpha_{\min}
+(\alpha_{\max}-\alpha_{\min})e^{-\Psi_t},\\
K_t &= \min\!\left(\lfloor\alpha_t N\rfloor,\,
|\mathcal{C}_t|\right),
\end{aligned}
\label{eq:adaptive_reuse_budget}
\end{equation}
here $N$ is the number of current patches, and $\alpha_{\min}$ and $\alpha_{\max}$ bound the target reuse fraction. The realized fraction is $K_t/N$. 
When fewer than $\lfloor\alpha_t N\rfloor$ patches are eligible, the cache reuses only those eligible patches, so realized reuse can fall below $\alpha_{\min}$.

The entropy values in Fig.~\ref{fig:spectral_entropy} illustrate the effect of the budget rule. Frames 12 and 66 have similar initial match fractions, but their entropies of 2.36 and 0.84 give $e^{-\Psi_t}$ values of 0.094 and 0.432, respectively. Frame 66 therefore receives a larger target budget. Whether it realizes that increase depends on the candidates that remain after alignment and refresh filtering.

\subsubsection{\textbf{Energy-Ranked Selection within Eligible Token Candidates}}
Given $K_t$, we rank candidates in $\mathcal{C}_t$ by ascending high-frequency energy and reuse the $K_t$ lowest-energy tokens from their aligned historical entries. All remaining patches are recomputed. The reused and recomputed initial visual-token features are merged in their original token order to update the cache for the next observation. Alg.~\ref{alg:freqcache} summarizes the process.

\begin{algorithm}[!t]
\caption{Joint Frequency--Token Caching Process}
\footnotesize
\label{alg:freqcache}
\begin{algorithmic}[1]
\Require Current observation $I_t$, previous observation $I_{t-1}$, cache $\mathcal{H}_{t-1}$
\Ensure Initial visual tokens and updated cache
\If{no valid history or cache}
    \State Compute all initial tokens, initialize cache; \Return
\EndIf
\State Compute spectral magnitude similarity
\If{similarity $<\tau_{\mathrm{mig}}$}
    \State Recompute all tokens and reset cache; \Return
\EndIf
\State Form preliminary same-position matches $\mathcal{P}_t$
\State Estimate the dominant displacement by phase correlation
\State Refine matches within the overlap and record the historical index of each aligned candidate
\State Compute local DCT high-frequency energy and refresh threshold
\State Retain in $\mathcal{C}_t$ each aligned candidate with $E_{i,j}\leq\theta_E^t$
\State Compute spectral entropy and reuse budget $K_t$ (Eq.~\eqref{eq:adaptive_reuse_budget})
\State Reuse the $K_t$ lowest-energy candidates; recompute remaining patches
\State Merge initial tokens and update cache
\end{algorithmic}
\end{algorithm}

\section{\textbf{Implementation}}
\label{tex:implementation}

\subsection{\textbf{GPU Workload and Resource Mapping}}
\label{tex:implementation_1}

\subsubsection{\textbf{Workload Decomposition and Kernel Mapping}}
\begin{figure}[!t]
    \centering
    \includegraphics[width=3.3in]{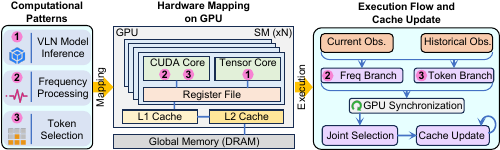}
    \vspace{-2mm}
    \caption{GPU Workload Mapping, Resource Organization, and Execution Flow.}
    \label{fig:6}
    \vspace{-4mm}
\end{figure}

As shown in Fig.~\ref{fig:6}, \textit{FreqCache} uses three GPU workloads: VLN inference, frequency processing, and patch/token selection, corresponding to matrix computation, spectral transforms, and data-intensive indexing. 
Implemented with PyTorch and CUDA, supported VLN matrix operators use Tensor Cores; frequency and patch/token operations use CUDA cores. 
cuFFT handles Fourier transforms, while custom CUDA kernels implement patch indexing, candidate selection, feature gathering, and token merging.

\subsubsection{\textbf{GPU Resource and Memory Organization}}
The workloads map to GPU SMs and the memory hierarchy. Registers and L1/L2 caches serve frequently accessed operands; global memory stores observations, parameters, and visual-token features. Keeping spectral cues and cached features GPU-resident reduces host--device transfers.

\subsection{\textbf{GPU Execution Flow and Cache Update}}
\label{tex:implementation_2}

\subsubsection{\textbf{Frequency--Token Workload Coordination}}
At each step, current and historical observations enter frequency and token branches. The frequency branch computes magnitude similarity, phase displacement, DCT high-frequency energy, and spectral entropy for alignment, refresh, and budgeting, respectively. The token branch matches current patches to historical initial visual features and prepares candidate indices. Both branches synchronize before joint selection.

\subsubsection{\textbf{Feature Reuse and Cache Write-Back}}
Joint selection supplies a current position $p$ and its historical source $\pi_t(p)$ for each cache hit. 
After matching, \textit{FreqCache} fills each of the $K_t$ hit slots $p$ with the cached embedding from $\pi_t(p)$ and uses current-frame embeddings for the remaining $N-K_t$ slots.
The merged buffer becomes the next cache, so consecutive hits can carry an embedding across multiple steps. All $N$ positions enter dense VLM prefill in current-frame order, preserving the projection and attention shapes. A displaced hit thus changes the source of the embedding while retaining its current spatial position.

\section{\textbf{Experiments}}
\label{tex:experiments}

\subsection{\textbf{Setup}}
We use DualVLN~\cite{wei2025dualvln} and StreamVLN~\cite{wei2025streamvln} to evaluate \textit{FreqCache} on R2R-CE~\cite{anderson2018vision,krantz2020beyond} and RxR-CE~\cite{ku2020rxr}.
For each model, we compare the original pipeline with two caching baselines (VLA-Cache~\cite{xu2025vlacache} and VLN-Cache~\cite{zheng2026vlncache}) and four pruning baselines (SparseVLM~\cite{zhang2025sparsevlm}, VisPruner~\cite{zhang2025vispruner}, FastVID~\cite{shen2025fastvid}, and the history-conditioned method of Wang et al.~\cite{wang2026history}).
We report navigation error (NE), oracle success (OS), success rate (SR), success weighted by path length (SPL), and normalized dynamic time warping (nDTW).
We report per-step inference latency on R2R-CE for an NVIDIA A100 and a Jetson Orin. Speedup is the ratio of original to accelerated latency on the same model and device.

\subsection{\textbf{Evaluation Results}}
\subsubsection{\textbf{Navigation performance and Latency}}
In Tabs.~\ref{tab:eval-dualvln} and~\ref{tab:eval-streamvln}, \textit{FreqCache} achieves the lowest per-step latency and the highest SR and SPL among the acceleration methods on both models.
On R2R-CE, it reduces A100 latency from 0.637 to 0.391 s for DualVLN and from 0.320 to 0.224 s for StreamVLN, giving 1.63$\times$ and 1.43$\times$ speedups, respectively. The corresponding Jetson Orin speedups are 1.64$\times$ and 1.44$\times$.
Across R2R-CE and RxR-CE, SR and SPL decrease by at most 0.7 percentage points relative to the original models.
Compared with VLN-Cache, the fastest baseline, \textit{FreqCache} lowers latency by a further 6.7--12.0\% and improves SPL by 0.5--0.8 points. It thus improves both sides of the latency--SPL trade-off at the evaluated settings.

\subsubsection{\textbf{Token Reuse during Navigation}}
\begin{figure*}[!t]
    \centering
    \vspace{-3mm}
    \includegraphics[width=6.7in]{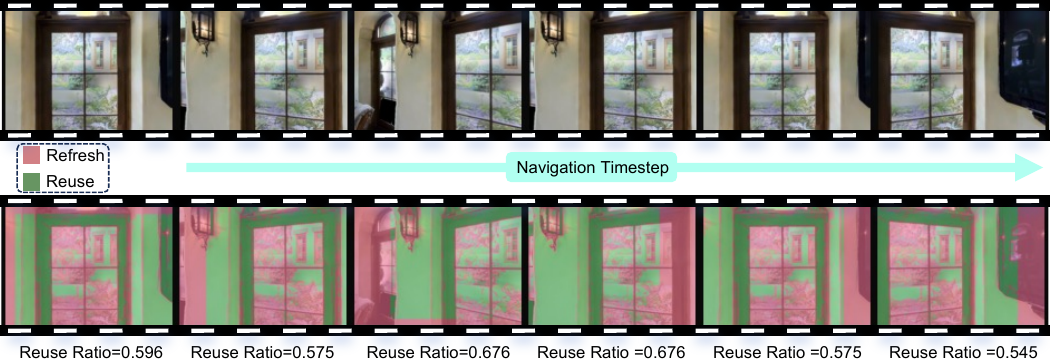}
    \vspace{-3mm}
    \caption{Token-Only and Frequency-Guided Reuse across 11 Navigation Views.}
    \label{fig:7}
    \vspace{-2mm}
\end{figure*}

Each column of Fig.~\ref{fig:7} compares two masks on the same view (green for reuse, pink for refresh). At the sharp turn, frequency-guided matching follows the shifted wall. In the third column, reuse rises from 3.3\% (token-only, top) to 46.9\% (frequency-guided, bottom). At the final turn, the lower mask refreshes the newly exposed edge while reusing 54.5\% of patches.

\begin{table}[t]
  \centering
  \vspace{-4mm}
  \caption{FreqCache (ours) and baselines on \textcolor{blue}{\textbf{DualVLN}}~\cite{wei2025dualvln}.}
  \vspace{-2mm}
  \label{tab:eval-dualvln}
  \scriptsize
  \setlength{\tabcolsep}{0.72pt}
  \renewcommand{\arraystretch}{0.98}
  \setlength{\aboverulesep}{0.2ex}
  \setlength{\belowrulesep}{0.2ex}
  \setlength{\cmidrulesep}{0.15ex}
  \begin{tabular*}{\columnwidth}{@{\extracolsep{\fill}}l*{4}{c}@{\hspace{3.5pt}}*{4}{c}@{\hspace{3.5pt}}cc@{}}
    \toprule
    \toprule
    \textbf{\multirow{2}{*}{Methods}} & \multicolumn{4}{c}{\textbf{R2R-CE}}
    & \multicolumn{4}{c}{\textbf{RxR-CE}}
    & \multicolumn{2}{c@{}}{\textbf{R2R Lat. (s)}} \\
    \cmidrule(lr){2-5}\cmidrule(lr){6-9}\cmidrule(l){10-11}
    ~
    & NE$\downarrow$ & OS$\uparrow$ & SR$\uparrow$ & SPL$\uparrow$
    & NE$\downarrow$ & SR$\uparrow$ & SPL$\uparrow$ & nDTW$\uparrow$
    & A100$\downarrow$ & Orin$\downarrow$ \\
    \midrule
    Original      & 4.05 & 70.7 & 64.3 & 58.5 & 4.58 & 61.4 & 51.8 & 70.0 & 0.637 & 4.20 \\
    VLA-Cache~\cite{xu2025vlacache} & 4.13 & 69.9 & 62.9 & 57.2 & 4.70 & 59.9 & 50.2 & 69.1 & 0.484 & 3.20 \\
    VLN-Cache~\cite{zheng2026vlncache} & 3.93 & 71.4 & 63.1 & 57.6 & 4.69 & 60.0 & 50.5 & 69.2 & 0.419 & 2.75 \\
    SparseVLM~\cite{zhang2025sparsevlm} & 4.16 & 69.5 & 62.8 & 57.0 & 4.74 & 59.4 & 49.7 & 68.8 & 0.548 & 3.70 \\
    VisPruner~\cite{zhang2025vispruner} & 4.35 & 66.6 & 61.2 & 56.3 & 4.89 & 59.2 & 50.1 & 68.5 & 0.584 & 3.90 \\
    FastVID~\cite{shen2025fastvid} & 4.31 & 67.5 & 61.9 & 56.9 & 4.85 & 59.6 & 50.5 & 68.7 & 0.567 & 3.78 \\
    Wang~\cite{wang2026history} & 4.26 & 68.3 & 62.5 & 57.5 & 4.82 & 60.0 & 50.9 & 69.0 & 0.502 & 3.44 \\
    \midrule[0.35pt]
    \rowcolor{gray!20}
    \textbf{\textit{FreqCache}}
                  & 3.90 & 71.6 & 63.8 & 58.1 & 4.63 & 60.8 & 51.3 & 69.8
                  & \textbf{0.391} & \textbf{2.56} \\
    \bottomrule
    \bottomrule
  \end{tabular*}
  \vspace{-4mm}
\end{table}

\subsection{\textbf{Ablation Study}}
In Tab.~\ref{tab:freqcache_ablation}, the caching variants differ in reuse ratio by at most 0.3 percentage points per model.
Relative to magnitude-only caching, full \textit{FreqCache} gains 1.7 SPL points on DualVLN and 1.6 on StreamVLN at a cost of 4 and 3 ms per step, respectively.
Removing phase alignment cuts SPL by 1.0 and 0.9 points, respectively, and removing high-frequency cues costs 0.7 points on each model.
At matched mean reuse (48.6\% on DualVLN and 46.5\% on StreamVLN), adaptive budgeting gains 0.4 SPL points over a fixed budget on each model.

\begin{figure}[!b]
    \centering
    \vspace{-4mm}
    \includegraphics[width=3.3in]{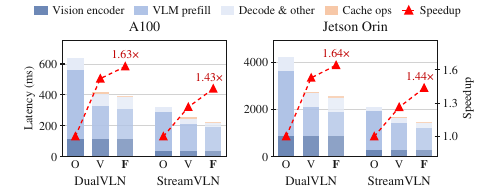}
    \vspace{-3mm}
    \caption{Per-Step Latency by Stage.}
    \label{fig:latency-breakdown}
\end{figure}

\subsection{\textbf{Discussion}}
\subsubsection{\textbf{Latency Breakdown}}
On DualVLN/A100, per-step latency is 637~ms for the original model (O), 419~ms for VLN-Cache (V), and 391~ms for \textit{FreqCache} (F). Of the 28~ms that \textit{FreqCache} saves over VLN-Cache, 23.6~ms comes from VLM prefill and 4.4~ms from cache operations (Fig.~\ref{fig:latency-breakdown}). Across 11 paired views (Fig.~\ref{fig:7}), \textit{FreqCache} refreshes 46.1\% of patches on average, compared with 90.7\% for token-only matching, while retaining all $N$ visual positions in dense prefill (Tab.~\ref{tab:freqcache-refresh-prefill}).

\subsubsection{\textbf{Reuse Budget}}
\begin{figure}[!b]
    \centering
    \vspace{-5mm}
    \includegraphics[width=3.3in]{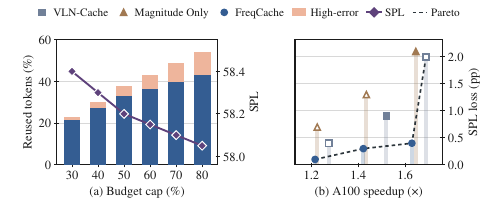}
    \vspace{-3mm}
    \caption{Cache Budget Sensitivity and Speed--Quality Trade-offs on R2R-CE.}
    \label{fig:reuse-budget}
\end{figure}

In Fig.~\ref{fig:reuse-budget}(a), bar height gives realized reuse, and the orange segment is the fraction of all patches reused with high error. Raising the cap from 30\% to 80\% increases realized reuse from about 23\% to 54\%, while SPL declines. In (b), the DualVLN settings in Table~\ref{tab:freqcache_ablation} yield similar A100 speedups for \textit{FreqCache} and magnitude-only caching ($1.63\times$ and $1.65\times$), with SPL losses of 0.4 and 2.1 points, respectively.

\begin{table}[t]
  \centering
  \vspace{-4mm}
  \caption{FreqCache (ours) and baselines on \textcolor{blue}{\textbf{StreamVLN}}~\cite{wei2025streamvln}.}
  \vspace{-2mm}
  \label{tab:eval-streamvln}
  \scriptsize
  \setlength{\tabcolsep}{0.72pt}
  \renewcommand{\arraystretch}{0.98}
  \setlength{\aboverulesep}{0.2ex}
  \setlength{\belowrulesep}{0.2ex}
  \setlength{\cmidrulesep}{0.15ex}
  \begin{tabular*}{\columnwidth}{@{\extracolsep{\fill}}l*{4}{c}@{\hspace{3.5pt}}*{4}{c}@{\hspace{3.5pt}}cc@{}}
    \toprule
    \toprule
    \textbf{\multirow{2}{*}{Methods}} & \multicolumn{4}{c}{\textbf{R2R-CE}}
    & \multicolumn{4}{c}{\textbf{RxR-CE}}
    & \multicolumn{2}{c@{}}{\textbf{R2R Lat. (s)}} \\
    \cmidrule(lr){2-5}\cmidrule(lr){6-9}\cmidrule(l){10-11}
    ~
    & NE$\downarrow$ & OS$\uparrow$ & SR$\uparrow$ & SPL$\uparrow$
    & NE$\downarrow$ & SR$\uparrow$ & SPL$\uparrow$ & nDTW$\uparrow$
    & A100$\downarrow$ & Orin$\downarrow$ \\
    \midrule
    Original      & 4.90 & 63.6 & 56.4 & 50.2 & 5.65 & 54.4 & 45.4 & 63.7 & 0.320 & 2.11 \\
    VLA-Cache~\cite{xu2025vlacache} & 4.98 & 62.9 & 55.2 & 49.2 & 5.78 & 53.0 & 44.1 & 62.9 & 0.266 & 1.76 \\
    VLN-Cache~\cite{zheng2026vlncache} & 4.96 & 63.1 & 55.4 & 49.3 & 5.78 & 53.0 & 44.1 & 62.9 & 0.253 & 1.67 \\
    SparseVLM~\cite{zhang2025sparsevlm} & 4.98 & 62.8 & 55.4 & 49.3 & 5.79 & 52.9 & 44.0 & 62.8 & 0.288 & 1.91 \\
    VisPruner~\cite{zhang2025vispruner} & 5.20 & 59.5 & 53.3 & 48.0 & 5.96 & 52.2 & 43.7 & 62.2 & 0.301 & 1.99 \\
    FastVID~\cite{shen2025fastvid} & 5.16 & 60.4 & 54.0 & 48.6 & 5.92 & 52.6 & 44.1 & 62.4 & 0.299 & 1.99 \\
    Wang~\cite{wang2026history} & 5.11 & 61.2 & 54.6 & 49.2 & 5.89 & 53.0 & 44.5 & 62.7 & 0.272 & 1.81 \\
    \midrule[0.35pt]
    \rowcolor{gray!20}
    \textbf{\textit{FreqCache}}
                  & 4.92 & 63.4 & 56.0 & 49.8 & 5.71 & 53.7 & 44.7 & 63.3
                  & \textbf{0.224} & \textbf{1.47} \\
    \bottomrule
    \bottomrule
  \end{tabular*}
  \vspace{-4mm}
\end{table}

\begin{table}[t]
  \centering
  \caption{Component ablations of FreqCache on R2R-CE.}
  \vspace{-2mm}
  \label{tab:freqcache_ablation}
  \scriptsize
  \setlength{\tabcolsep}{1.25pt}
  \renewcommand{\arraystretch}{0.98}
  \setlength{\aboverulesep}{0.2ex}
  \setlength{\belowrulesep}{0.2ex}
  \setlength{\cmidrulesep}{0.15ex}
  \begin{tabular*}{\columnwidth}{@{\extracolsep{\fill}}l*{3}{c}@{\hspace{6pt}}*{3}{c}@{}}
    \toprule
    \toprule
    \textbf{\multirow{2}{*}{Configuration}} & \multicolumn{3}{c}{\textbf{DualVLN~\cite{wei2025dualvln}}}
    & \multicolumn{3}{c@{}}{\textbf{StreamVLN~\cite{wei2025streamvln}}} \\
    \cmidrule(lr){2-4}\cmidrule(l){5-7}
    ~
    & SPL$\uparrow$ & Reuse (\%) & Lat. (s)$\downarrow$
    & SPL$\uparrow$ & Reuse (\%) & Lat. (s)$\downarrow$ \\
    \midrule
    Original
      & \textbf{58.5} & 0.0 & 0.637
      & \textbf{50.2} & 0.0 & 0.320 \\
    \midrule
    Magnitude Only
      & 56.4 & 48.5 & \textbf{0.387}
      & 48.2 & 46.4 & \textbf{0.221} \\
    w/o Phase Alignment
      & 57.1 & 48.8 & 0.388
      & 48.9 & 46.6 & 0.222 \\
    w/o HF Cues
      & 57.4 & 48.7 & 0.388
      & 49.1 & 46.7 & 0.222 \\
    Fixed Reuse Budget
      & 57.7 & 48.6 & 0.390
      & 49.4 & 46.5 & 0.223 \\
    \midrule
    \textit{Full \textbf{FreqCache}}
      & 58.1 & 48.6 & 0.391
      & 49.8 & 46.5 & 0.224 \\
    \bottomrule
    \bottomrule
  \end{tabular*}
  \vspace{-4mm}
\end{table}

\begin{table}[!t]
    \centering
    \caption{Patch Refresh and A100 Latency with \textit{FreqCache}}
    \vspace{-2mm}
    \label{tab:freqcache-refresh-prefill}
    \scriptsize
    \setlength{\tabcolsep}{4.5pt}
    \renewcommand{\arraystretch}{0.85}
    \resizebox{0.98\columnwidth}{!}{%
        \begin{tabular}{@{}lrrrlrrr@{}}
            \toprule
            \toprule
            \multicolumn{4}{c}{\textbf{Patch refresh (\%)}} &
            \multicolumn{4}{c}{\textbf{DualVLN/A100 (ms)}} \\
            \cmidrule(lr){1-4}\cmidrule(l){5-8}
            Method & Mean & Med. & Max. &
            Method & Prefill & Cache & Step \\
            \midrule
            Token-only & 90.7 & 89.8 & 97.8 &
            VLN-Cache  & 214.0 & 12.0 & 419 \\
            \textit{\textbf{FreqCache}} & 46.1 & 45.5 & 59.8 &
            \textit{\textbf{FreqCache}} & 190.4 & 7.6 & 391 \\
            \bottomrule
            \bottomrule
        \end{tabular}%
    }
    \vspace{-4mm}
\end{table}

\section{\textbf{Conclusion}}
\label{tex:conclusion}

\textit{FreqCache} uses spectral cues to recover shifted patch matches, refresh structurally sensitive patches, and adapt reuse to each frame. All visual positions enter dense prefill in their original order. On R2R-CE, it accelerates per-step inference by $1.43$--$1.64\times$ across two models and two GPUs. At the same mean reuse ratio, adaptive budgeting improves SPL by 0.4 points over a fixed budget on both models.

\clearpage
\balance
\bibliographystyle{IEEEtran}
\bibliography{ref/reference}

@article{streamvln,
  title={Streamvln: Streaming vision-and-language navigation via slowfast context modeling},
  author={Wei, Meng and Wan, Chenyang and Yu, Xiqian and Wang, Tai and Yang, Yuqiang and Mao, Xiaohan and Zhu, Chenming and Cai, Wenzhe and Wang, Hanqing and Chen, Yilun and others},
  journal={arXiv preprint arXiv:2507.05240},
  year={2025}
}

@article{pertsch2025fast,
  title={Fast: Efficient action tokenization for vision-language-action models},
  author={Pertsch, Karl and Stachowicz, Kyle and Ichter, Brian and Driess, Danny and Nair, Suraj and Vuong, Quan and Mees, Oier and Finn, Chelsea and Levine, Sergey},
  journal={arXiv preprint arXiv:2501.09747},
  year={2025}
}

@misc{internvla-n1,
    title = {{InternVLA-N1: An} Open Dual-System Navigation Foundation Model with Learned Latent Plans},
    author = {InternNav Team},
    year = {2025},
    booktitle={arXiv},
}

@inproceedings{chen2024mapgpt,
  title={Mapgpt: Map-guided prompting with adaptive path planning for vision-and-language navigation},
  author={Chen, Jiaqi and Lin, Bingqian and Xu, Ran and Chai, Zhenhua and Liang, Xiaodan and Wong, Kwan-Yee},
  booktitle={Proceedings of the 62nd Annual Meeting of the Association for Computational Linguistics (Volume 1: Long Papers)},
  pages={9796--9810},
  year={2024}
}

@article{zhang2024navid,
  title={Navid: Video-based vlm plans the next step for vision-and-language navigation},
  author={Zhang, Jiazhao and Wang, Kunyu and Xu, Rongtao and Zhou, Gengze and Hong, Yicong and Fang, Xiaomeng and Wu, Qi and Zhang, Zhizheng and Wang, He},
  journal={arXiv preprint arXiv:2402.15852},
  year={2024}
}

@article{long2024instructnav,
  title={Instructnav: Zero-shot system for generic instruction navigation in unexplored environment},
  author={Long, Yuxing and Cai, Wenzhe and Wang, Hongcheng and Zhan, Guanqi and Dong, Hao},
  journal={arXiv preprint arXiv:2406.04882},
  year={2024}
}

@inproceedings{kang2025harnessing,
  title={Harnessing Input-Adaptive Inference for Efficient VLN},
  author={Kang, Dongwoo and Perincherry, Akhil and Coalson, Zachary and Gabriel, Aiden and Lee, Stefan and Hong, Sanghyun},
  booktitle={Proceedings of the IEEE/CVF International Conference on Computer Vision},
  pages={8219--8229},
  year={2025}
}

@inproceedings{anderson2018vision,
  title={Vision-and-language navigation: Interpreting visually-grounded navigation instructions in real environments},
  author={Anderson, Peter and Wu, Qi and Teney, Damien and Bruce, Jake and Johnson, Mark and S{\"u}nderhauf, Niko and Reid, Ian and Gould, Stephen and Van Den Hengel, Anton},
  booktitle={Proceedings of the IEEE conference on computer vision and pattern recognition},
  pages={3674--3683},
  year={2018}
}

@article{xu2025vlacache,
  title   = {VLA-Cache: Efficient Vision-Language-Action Manipulation via Adaptive Token Caching},
  author  = {Xu, Siyu and Wang, Yunke and Xia, Chenghao and Zhu, Dihao and Huang, Tao and Xu, Chang},
  journal = {arXiv preprint arXiv:2502.02175},
  year    = {2025}
}

@inproceedings{krantz2020beyond,
  title     = {Beyond the Nav-Graph: Vision-and-Language Navigation in Continuous Environments},
  author    = {Krantz, Jacob and Wijmans, Erik and Majumdar, Arjun and Batra, Dhruv and Lee, Stefan},
  booktitle = {European Conference on Computer Vision (ECCV)},
  year      = {2020}
}

@inproceedings{zhang2025uninavid,
  title={Uni-NaVid: A Video-based Vision-Language-Action Model for Unifying Embodied Navigation Tasks},
  author={Zhang, Jiazhao and Wang, Kunyu and Wang, Shaoan and Li, Minghan and Liu, Haoran and Wei, Songlin and Wang, Zhongyuan and Zhang, Zhizheng and Wang, He},
  booktitle={Robotics: Science and Systems (RSS)},
  year={2025},
  url={https://arxiv.org/abs/2412.06224}
}

@article{wei2025streamvln,
  title={StreamVLN: Streaming Vision-and-Language Navigation via SlowFast Context Modeling},
  author={Wei, Meng and Wan, Chenyang and Yu, Xiqian and Wang, Tai and Yang, Yuqiang and Mao, Xiaohan and Zhu, Chenming and Cai, Wenzhe and Wang, Hanqing and Chen, Yilun and Liu, Xihui and Pang, Jiangmiao},
  journal={arXiv preprint arXiv:2507.05240},
  year={2025}
}

@article{wang2025fouriervlm,
  title={Fourier-VLM: Compressing Vision Tokens in the Frequency Domain for Large Vision-Language Models},
  author={Wang, Huanyu and Kai, Jushi and Bai, Haoli and Hou, Lu and Jiang, Bo and He, Ziwei and Lin, Zhouhan},
  journal={arXiv preprint arXiv:2508.06038},
  year={2025},
  url={https://arxiv.org/abs/2508.06038}
}

@article{liu2025freqca,
  title={FreqCa: Accelerating Diffusion Models via Frequency-Aware Caching},
  author={Liu, Jiacheng and Cai, Peiliang and Zhou, Qinming and Lin, Yuqi and Kong, Deyang and Huang, Benhao and Pan, Yupei and Xu, Haowen and Zou, Chang and Tang, Junshu and Zheng, Shikang and Zhang, Linfeng},
  journal={arXiv preprint arXiv:2510.08669},
  year={2025},
  url={https://arxiv.org/abs/2510.08669}
}

@article{zheng2026vlncache,
  title={VLN-Cache: Enabling Token Caching for VLN Models with Visual/Semantic Dynamics Awareness},
  author={Zheng, Zihao and Mao, Zhihao and Zhou, Xingyue and Chen, Jiayu and Li, Maoliang and Sun, Xinhao and Zou, Hailong and Zhang, Zhaobo and Liu, Xuanzhe and Cao, Donggang and Mei, Hong and Chen, Xiang},
  journal={arXiv preprint arXiv:2603.07080},
  year={2026},
  url={https://arxiv.org/abs/2603.07080}
}

@inproceedings{yang2025freqts,
  title={FreqTS: Frequency-Aware Token Selection for Accelerating Diffusion Models},
  author={Yang, Xinye and Yang, Yuxin and Pang, Haoran and Tian, Aaron Xuxiang and Li, Luking},
  booktitle={Proceedings of the AAAI Conference on Artificial Intelligence},
  volume={39},
  number={9},
  pages={9310--9317},
  year={2025},
  doi={10.1609/aaai.v39i9.33008}
}

@inproceedings{kai2026freqkv,
  title={FreqKV: Key-Value Compression in Frequency Domain for Context Window Extension},
  author={Kai, Jushi and Wang, Yixuan and Zeng, Boyi and Bai, Haoli and Jiang, Bo and He, Ziwei and Lin, Zhouhan},
  booktitle={International Conference on Learning Representations},
  year={2026}
}

@inproceedings{yang2026flashcache,
  title={Revisiting Multimodal KV Cache Compression: A Frequency-Domain-Guided Outlier-KV-Aware Approach},
  author={Yang, Yaoxin and Ye, Peng and Tan, Xudong and Tu, Chongjun and Zhao, Maosen and Hao, Jia and Chen, Tao},
  booktitle={Proceedings of the IEEE/CVF Conference on Computer Vision and Pattern Recognition},
  pages={39550--39560},
  year={2026}
}

@article{wei2025dualvln,
  title={Ground Slow, Move Fast: A Dual-System Foundation Model for Generalizable Vision-and-Language Navigation},
  author={Wei, Meng and Wan, Chenyang and Peng, Jiaqi and Yu, Xiqian and Yang, Yuqiang and Feng, Delin and Cai, Wenzhe and Zhu, Chenming and Wang, Tai and Pang, Jiangmiao and Liu, Xihui},
  journal={arXiv preprint arXiv:2512.08186},
  year={2025}
}

@inproceedings{ku2020rxr,
  title={Room-Across-Room: Multilingual Vision-and-Language Navigation with Dense Spatiotemporal Grounding},
  author={Ku, Alexander and Anderson, Peter and Patel, Roma and Ie, Eugene and Baldridge, Jason},
  booktitle={Proceedings of the 2020 Conference on Empirical Methods in Natural Language Processing},
  year={2020},
  doi={10.18653/v1/2020.emnlp-main.356}
}

@inproceedings{zhang2025sparsevlm,
  title={{SparseVLM}: Visual Token Sparsification for Efficient Vision-Language Model Inference},
  author={Zhang, Yuan and Fan, Chun-Kai and Ma, Junpeng and Zheng, Wenzhao and Huang, Tao and Cheng, Kuan and Gudovskiy, Denis A. and Okuno, Tomoyuki and Nakata, Yohei and Keutzer, Kurt and Zhang, Shanghang},
  booktitle={Proceedings of the 42nd International Conference on Machine Learning},
  pages={74840--74857},
  year={2025}
}

@inproceedings{zhang2025vispruner,
  title={Beyond Text-Visual Attention: Exploiting Visual Cues for Effective Token Pruning in {VLMs}},
  author={Zhang, Qizhe and Cheng, Aosong and Lu, Ming and Zhang, Renrui and Zhuo, Zhiyong and Cao, Jiajun and Guo, Shaobo and She, Qi and Zhang, Shanghang},
  booktitle={Proceedings of the IEEE/CVF International Conference on Computer Vision},
  pages={20857--20867},
  year={2025}
}

@inproceedings{shen2025fastvid,
  title={{FastVID}: Dynamic Density Pruning for Fast Video Large Language Models},
  author={Shen, Leqi and Gong, Guoqiang and He, Tao and Zhang, Yifeng and Liu, Pengzhang and Zhao, Sicheng and Ding, Guiguang},
  booktitle={Advances in Neural Information Processing Systems},
  volume={38},
  year={2025}
}

@inproceedings{wang2026history,
  author={Wang, Qitong and Liang, Yijun and Li, Ming and Zhou, Tianyi and Rasmussen, Christopher},
  title={History-Conditioned Spatio-Temporal Visual Token Pruning for Efficient Vision-Language Navigation},
  booktitle={IEEE/RSJ International Conference on Intelligent Robots and Systems (IROS)},
  year={2026}
}

@article{li2025egoprune,
  author  = {Li, Jiaao and Li, Kaiyuan and Gao, Chen and Li, Yong and Chen, Xinlei},
  title   = {{EgoPrune}: Efficient Token Pruning for Egomotion Video Reasoning in Embodied Agent},
  journal = {arXiv preprint arXiv:2507.15428},
  year    = {2025},
  doi     = {10.48550/arXiv.2507.15428}
}

\end{document}